\documentclass[11pt,a4paper]{article}

\usepackage[utf8]{inputenc}
\usepackage[english]{babel}
\usepackage{amsmath, amsfonts, amssymb}
\usepackage{graphicx}
\usepackage{booktabs}
\usepackage{geometry}
\usepackage{hyperref}
\usepackage{xcolor}
\usepackage{caption}
\usepackage{fancyhdr}
\usepackage{cite} 
\usepackage{authblk} 

\makeatletter
\renewcommand\AB@authnote[1]{} 
\renewcommand\AB@affilnote[1]{} 
\makeatother

\title{\textbf{Neural-Actuarial Longevity Forecasting: \\ Anchoring LSTMs for Explainable Risk Management}}

\author{
    \textbf{Davide Rindori} \\
    \textit{SAV Actuarial Candidate, ETH Zürich} \\
    \textit{PhD in Physics, University of Florence (2021)} \\
    \texttt{drindori@student.ethz.ch} \quad | \quad \texttt{rindori.d@gmail.com}
}

\date{\today}

\begin{document}
\maketitle

\begin{abstract}
Traditional multi-population models, such as the Li-Lee framework, rely on the assumption of mean-reverting country-specific deviations. However, recent data from high-longevity clusters suggest a systemic break in this paradigm. We identify a \textit{stationarity paradox} where mortality residuals in countries like Sweden and West Germany exhibit persistent unit roots, leading to a systematic mispricing of longevity risk in linear models.

To address these non-linearities, we propose \textit{Hybrid-Lift}, a neural-actuarial framework that combines Hierarchical LSTM networks with a Mean-Bias Correction (MBC) anchoring mechanism. Positioned as a governance-friendly model challenger rather than a replacement of classical approaches, the framework exhibits selective superiority on out-of-sample validation (2012--2020): it outperforms Li-Lee by 17.40\% in Sweden and 12.57\% in West Germany, while remaining comparable for near-linear regimes such as Switzerland and Japan.

We complement the predictive model with an integrated governance suite comprising SHAP-based cross-country influence mapping, a dual uncertainty framework for regulatory capital calibration (Swiss ES 99.0\% of +1.153 years), and a reverse stress test identifying the critical shock threshold for solvency buffer exhaustion. This research provides evidence that neural networks, when properly anchored by actuarial principles, can serve as effective model challengers for longevity risk management under the SST and Solvency II standards.

\vspace{0.4cm}
\noindent \textbf{Keywords:} Mortality Forecasting, Multi-population Longevity, Neural-Actuarial Modeling, Model Challenger, Explainable AI (XAI), Swiss Solvency Test (SST).

\vspace{0.2cm}
\noindent The complete codebase, including data processing, model training, and validation notebooks, is publicly available.\footnote{\url{https://github.com/davide-rindori/Actuarial-DS-Portfolio/tree/main/04_Multi_Population_Longevity_XAI}}
\end{abstract}

\section{Introduction}

The systemic increase in human life expectancy presents one of the most significant challenges for the modern insurance industry and social security systems. For institutional reinsurers like Swiss Re, the accurate quantification of longevity risk is not merely a pricing exercise but a fundamental requirement for capital adequacy under the Swiss Solvency Test (SST) and Solvency II frameworks \cite{swissre2022}. Traditional stochastic mortality models, primarily the Lee-Carter framework \cite{leecarter1992} and its multi-population extension by Li and Lee \cite{lilee2005}, have long served as the industry benchmark. These models rely on the assumption of a common mortality trend shared across coherent populations, where individual deviations are assumed to be mean-reverting and stationary.

However, recent demographic data from high-longevity clusters (1956--2020) suggest a structural shift. The mortality improvement deceleration observed post-2011 has exposed a \textit{stationarity paradox}: for core nations like Sweden, West Germany, and the Netherlands, country-specific residuals exhibit persistent unit roots rather than mean reversion, while Switzerland displays an inertial drift that defies simple classification. By forcing these persistent drifts back to a common mean, traditional linear models may systematically underestimate local longevity trends, leading to a miscalibration of Solvency Capital Requirements (SCR).

The emergence of actuarial machine learning, pioneered by Wüthrich and Merz \cite{wuthrich2021statistical, wuthrich2023}, provides a path to transcend these linear constraints. While Recurrent Neural Networks (RNNs) and Long Short-Term Memory (LSTM) architectures are theoretically capable of capturing non-linear temporal dependencies, their adoption in internal risk models has been hindered by two main factors: the black-box nature of Deep Learning and the risk of unconstrained integration drift in long-term projections.

In this work, we introduce \textit{Hybrid-Lift}, a neural-actuarial framework designed to bridge the gap between algorithmic flexibility and actuarial governance. Our approach differs from existing literature in three key dimensions:
\begin{itemize}
    \item \textbf{Neural-Actuarial Anchoring}: Instead of modeling absolute levels, we utilize an LSTM to predict first-differences ($\Delta K_t$), coupled with a Mean-Bias Correction (MBC) mechanism to ensure long-term stability and biological consistency.
    \item \textbf{Explainable Governance}: We move beyond point estimates by integrating SHapley Additive exPlanations (SHAP) to map cross-country influences. Following the methodological framework for actuarial interpretability proposed in \cite{shapforactuaries2023}, we open the black box of the LSTM to identify lead-lag relationships within the cluster, revealing a structured influence hierarchy among frontier nations, ensuring the model remains auditable for regulatory purposes.
    \item \textbf{Uncertainty Quantification}: We implement a dual uncertainty framework that combines Monte Carlo Dropout \cite{gal2016dropout} for model uncertainty with a process noise component calibrated on the historical variability of the Li-Lee factor structure, allowing for direct calibration of Value-at-Risk (VaR) and Expected Shortfall (ES) that reflects both epistemic and process uncertainty.
\end{itemize}

A key question motivating this research is whether a Hierarchical LSTM can implicitly learn the common factor structure without explicit SVD decomposition. Our results indicate that the framework achieves selective superiority: improvements of 17.40\% (Sweden) and 12.57\% (West Germany) in regimes where persistent unit roots invalidate the Li-Lee mean-reversion assumption, while remaining comparable to the linear benchmark for near-linear trajectories such as Switzerland and Japan. The model also reveals limitations: it exhibits a conservative bias during extreme volatility (e.g., the 2020 pandemic shock) and requires a minimum 10-year lookback window to outperform Li-Lee. These findings position Hybrid-Lift not as a replacement for classical actuarial models, but as a governance-friendly model challenger: a neural overlay that adds value precisely where linear assumptions break down, while providing an integrated explainability and capital calibration pipeline that classical models lack.

By focusing on a Frontier Cluster of six high-longevity nations, this paper provides evidence that neural networks, when properly anchored by actuarial principles, can complement and extend classical models for longevity risk management in both academic and industrial applications. The stochastic trajectories produced by the framework are directly applicable to the pricing of longevity-linked financial instruments such as longevity swaps, as illustrated in the accompanying codebase.

\subsection{Related Work}
The application of neural networks to mortality modeling has gained significant traction in recent years. Richman and Wüthrich \cite{richman2021} proposed a feed-forward network for Lee-Carter parameter estimation, showing that neural architectures can replicate and extend classical actuarial decompositions. Perla et al. \cite{perla2021} introduced a recurrent framework for multi-population mortality, showing that LSTMs can capture cross-country dependencies without explicit factor decomposition. Nigri et al. \cite{nigri2019} applied deep learning to single-population mortality forecasting, highlighting the potential of recurrent architectures for long-horizon projections. More recently, Robben, Antonio, and Kleinow \cite{robben2025} developed advanced statistical methods for fine-grained mortality data, showing that modern actuarial applications increasingly require flexible, data-driven approaches that go beyond classical parametric assumptions.

Our work builds upon these contributions but differs in four respects. First, we explicitly model first-differences rather than absolute levels, addressing the non-stationarity issue that limits the generalization of level-based neural approaches. Second, we introduce the Mean-Bias Correction (MBC) mechanism to anchor neural predictions to actuarial baselines, a step that is absent in prior neural mortality literature. Third, we integrate SHAP-based explainability following the framework proposed by Mayer, Meier, and Wüthrich \cite{shapforactuaries2023}, providing a governance layer that is essential for regulatory acceptance of internal models under SST and Solvency II. Fourth, we decompose the total forecast uncertainty into model uncertainty (via Monte Carlo Dropout) and process uncertainty (calibrated on the Li-Lee historical variability), a distinction that is standard in probabilistic machine learning but has not been explicitly operationalized in prior neural mortality frameworks for regulatory capital calculations.

\section{Data and Cluster Selection}

\subsection{The Frontier Mortality Cluster}
The empirical foundation of this study is the Frontier Mortality Cluster, which includes Switzerland (CHE), Sweden (SWE), Norway (NOR), West Germany (DEUTW), the Netherlands (NLD), and Japan (JPN). Data were retrieved from the Human Mortality Database\footnote{\url{https://www.mortality.org}} (HMD) \cite{hmd2023} for the period 1956--2020.

The selection of these specific $N=6$ populations is not arbitrary but follows a deliberate signal-focused approach. In multi-population modeling, there is an intrinsic trade-off between statistical volume and trend coherence. While increasing $N$ (e.g., adding the USA, UK, or Eastern European countries) would provide more data points, it would introduce structural heterogeneity (distinct health shocks, different socio-economic safety nets, and divergent medical standards) that contaminate the frontier mortality signal. 

We selected this cluster based on three criteria:
\begin{enumerate}
    \item \textbf{Historical Integrity}: These nations possess the most reliable long-term records in the HMD, minimizing reporting bias.
    \item \textbf{Socio-Economic Homogeneity}: These countries share comparable longevity outcomes, advanced medical infrastructure, and high socio-economic development, forming a coherent demographic frontier.
    \item \textbf{Mathematical Balance}: A common time window is a prerequisite for the Singular Value Decomposition (SVD) in the Li-Lee framework \cite{lilee2005} to extract a balanced common factor.
\end{enumerate}

Each population serves a distinct functional role within the cluster: Switzerland and West Germany provide the DACH regional anchor; Sweden and Norway contribute the Nordic dimension, with Sweden offering the longest high-quality HMD series and Norway serving as a small-population robustness test; the Netherlands represents the global benchmark in pension and longevity research; and Japan provides the extra-European biological frontier signal. We use West Germany (DEUTW) rather than unified Germany to avoid the structural discontinuity introduced by the 1990 reunification, which could contaminate the mortality trend with a compositional shift unrelated to biological longevity dynamics. Alternative candidates such as Denmark, Austria, or France were considered but excluded: Denmark and Austria would introduce redundancy with the existing Nordic and DACH components without adding informational diversity, while France and other Mediterranean populations would weaken the cluster's socio-economic coherence.

If we were to increase the number of countries, we would expect a trend dilution effect: the common factor $K_t$ would lose its ability to represent the leading edge of longevity, potentially masking the non-linearities we aim to capture. Conversely, a cluster of $N=6$ is sufficiently large to diversify local volatility (e.g., Norway's small-sample noise) while remaining focused enough to define a clear, shared biological trajectory.

\subsection{Temporal Horizon and Data Availability}
The dataset spans from 1956 to 2020. While certain HMD populations, most notably Sweden, provide records dating back to 1751, the choice of 1956 is a constraint determined by the synchronized availability of high-quality data for West Germany and Japan. Specifically, Japanese mortality records achieve full reliability and synchronization with European standards in the HMD only in the post-WWII period. 

\begin{figure}[!htbp]
    \centering
    \includegraphics[width=0.8\textwidth]{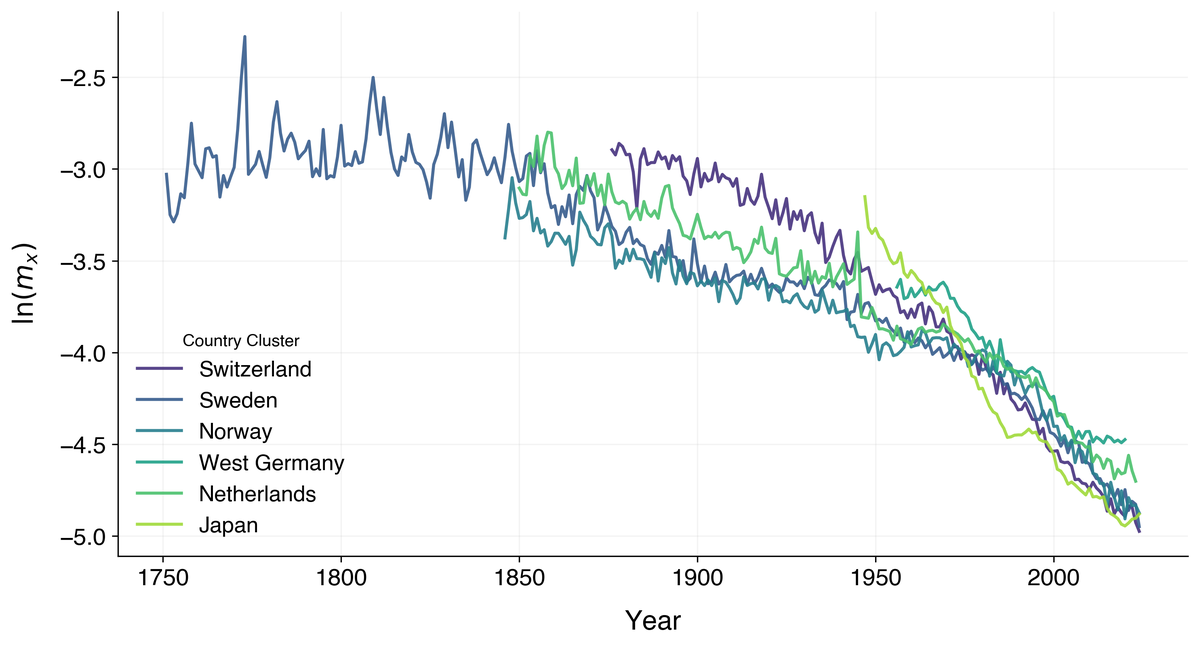}
    \caption{Log-mortality rates at age 65 across the frontier cluster (1956--2020). The convergence of trajectories justifies the multi-population approach.}
    \label{fig:comparison_65}
\end{figure}

Including a longer historical horizon would necessitate the exclusion of these two important nations. From a modeling perspective, the loss of the Japanese frontier signal and the German demographic weight would outweigh the benefits of a longer time series. Furthermore, the post-1950 era represents a structurally distinct regime characterized by the "cardiovascular revolution" \cite{mesle2002} and modernized medical protocols. As illustrated in Figure \ref{fig:comparison_65}, this window captures the significant convergence of mortality rates at advanced ages across the cluster, providing a robust baseline for multi-population training.

\subsection{Switzerland as the Primary Case Study}
While the \textit{Hybrid-Lift} framework is trained on the full six-country cluster to leverage multi-population correlations and shared biological trends, this research designates Switzerland as the primary case study for results validation and regulatory scenario analysis. This choice is motivated by the requirements of the Swiss Solvency Test (SST) and the specific needs of the Zurich-based actuarial ecosystem. Consequently, while numerical benchmarks are computed for all populations to ensure model robustness, the detailed analysis of explainability (XAI), stochastic fan charts, and capital requirements will focus specifically on the Swiss demographic trajectory.

\subsection{Mathematical Framework and Preprocessing}
The fundamental observable is the Central Death Rate, defined as:
\begin{equation}
    m_{x,t,i} = \frac{D_{x,t,i}}{E_{x,t,i}}
\end{equation}
where $D_{x,t,i}$ represents the number of deaths and $E_{x,t,i}$ the exposure at risk for age $x \in [0, 90]$, year $t \in [1956, 2020]$, and country $i$. All rates are computed for both sexes combined, consistent with the standard multi-population framework of Li and Lee \cite{lilee2005}. We operate in the log-domain:
\begin{equation}
    y_{x,t,i} = \ln(m_{x,t,i} + \epsilon)
\end{equation}
where $\epsilon = 10^{-10}$ is a small constant ensuring numerical stability for cells with zero deaths (particularly relevant for the Norwegian population). The logarithmic transformation is justified by the Gompertz law, linearizing the exponential growth of mortality with age, and guarantees that back-transformed rates remain strictly positive. We truncate the age range at $x \le 90$ to eliminate the high-frequency noise and erratic volatility typical of "oldest-old" data.

\subsection{The Stationarity Paradox}\label{sec:stationarity_paradox}
The Li-Lee framework assumes that country-specific deviations $k_{t,i}$ from the common trend $K_t$ follow a stationary, mean-reverting process. Mathematically, this implies that the residuals can be modeled as a first-order autoregressive process, $AR(1)$:
\begin{equation}
    k_{t,i} = \phi_i k_{t-1,i} + \xi_{t,i}
\end{equation}
where $\phi_i$ is the autoregressive coefficient governing the speed of mean reversion for country $i$, with $|\phi_i| < 1$ ensuring stationarity, and $\xi_{t,i} \sim \mathcal{N}(0, \sigma_{\xi}^2)$ is white noise. In this regime, any local deviation from the common mortality trend is transitory, and the series eventually returns to its mean over time. To verify this fundamental assumption, we performed a conflict analysis using the Augmented Dickey-Fuller (ADF) test \cite{dickfFuller1979}, which tests for the presence of a unit root, and the Kwiatkowski-Phillips-Schmidt-Shin (KPSS) test \cite{kpss1992}, which assesses the null hypothesis of stationarity.

\begin{table}[!htbp]
\centering
\caption{Stationarity Analysis of Country-Specific Residuals (1956--2020)}
\label{tab:stationarity}
\begin{tabular}{@{}lrrl@{}}
\toprule
Country & ADF ($p$-value) & KPSS ($p$-value) & Interpretation \\ \midrule
Norway & 0.0000 & 0.0889 & Stationary (Both PASS) \\
Japan & 0.0424 & 0.0165 & Persistent Drift (Conflict) \\
Netherlands & 0.1038 & 0.0100 & Unit Root (Both FAIL) \\
Switzerland & 0.7661 & 0.1000 & Inertial (Conflict) \\
Sweden & 0.9289 & 0.0100 & Unit Root (Both FAIL) \\
West Germany & 0.9467 & 0.0100 & Unit Root (Both FAIL) \\ \bottomrule
\end{tabular}
\vspace{0.2cm}

\footnotesize{\textit{Note:} ADF tests $H_0$: series has a unit root (PASS if $p < 0.05$). KPSS tests $H_0$: series is stationary (PASS if $p > 0.05$). Both PASS indicates robust stationarity; Both FAIL indicates unit root evidence; Conflict indicates persistent structural drift.}
\end{table}

\begin{figure}[!htbp]
    \centering
    \includegraphics[width=0.85\textwidth]{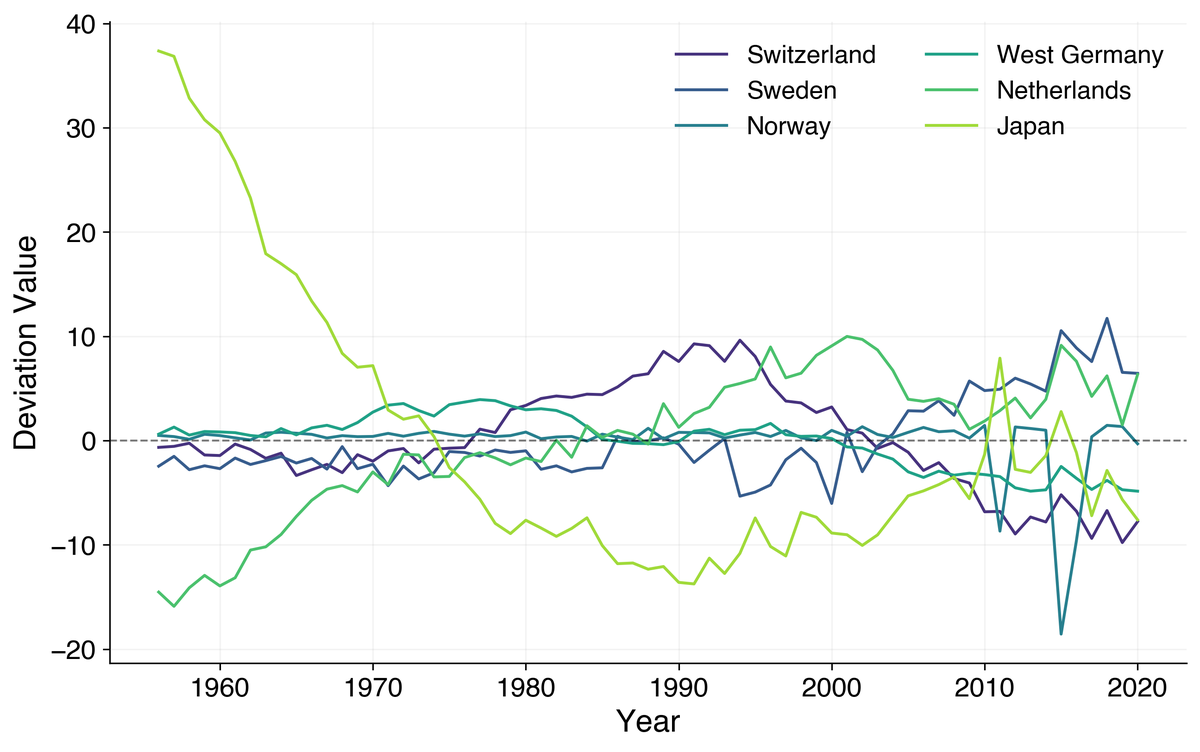}
    \caption{Evolution of country-specific factors $k_{t,i}$ for the frontier cluster. The trajectories of Switzerland and Sweden show persistent deviations from zero, whereas Norway's volatility leads to frequent mean-crossings.}
    \label{fig:specific_kt}
\end{figure}

The results in Table \ref{tab:stationarity} expose the \textit{stationarity paradox}: for core European populations (Sweden, West Germany, and the Netherlands), both tests agree on the presence of unit roots ($\phi_i \approx 1$), while Switzerland and Japan exhibit conflicting signals, suggesting persistent structural drifts that are neither clearly stationary nor clearly non-stationary. This divergence highlights a fundamental distinction between visual stability and stochastic reactivity, as illustrated by the evolution of specific factors in Figure \ref{fig:specific_kt}. Paradoxically, high-volatility populations like Norway pass the stationarity tests not because they are more stable, but because their frequent oscillations cross the mean with sufficient momentum for the ADF test to detect a reversion signal. 

Conversely, the Inertia Problem observed in core European countries reveals that mortality residuals behave like a Random Walk. While these series appear visually smooth, they lack a corrective force: any shock to the mortality trend becomes a permanent structural shift. Consequently, these populations exhibit a long-term divergence from the common cluster trend through persistent local drifts. The failure of the stationarity assumption in linear actuarial models is the primary methodological driver for our transition to LSTM-based modeling, which is natively designed to internalize non-stationary sequences and long-term temporal dependencies without forcing artificial mean-reversion.

\section{Methodology: From Actuarial Baselines to Neural Networks}

\subsection{The Li-Lee Benchmark Model}
The primary actuarial benchmark for multi-population mortality is the Li-Lee framework \cite{lilee2005}, which extends the Lee-Carter model \cite{leecarter1992} by imposing a common factor structure. Both models belong to the generalized age-period-cohort (GAPC) family formalized by Villegas et al. \cite{villegas2018stmomo}, which provides a unifying statistical framework for the vast majority of stochastic mortality projection models proposed to date. The fundamental assumption is that coherent populations share a long-term mortality trend, while individual country-specific deviations remain stationary.

The log-mortality rate for age $x$, year $t$, and country $i$ is decomposed as follows:
\begin{equation}
    y_{x,t,i} = \alpha_{x,i} + B_x K_t + b_{x,i} k_{t,i} + \varepsilon_{x,t,i}
\end{equation}
where:
\begin{itemize}
    \item $\alpha_{x,i}$ represents the average log-mortality profile for country $i$ at age $x$.
    \item $K_t$ is the common mortality index representing the shared trend of the entire cluster.
    \item $B_x$ is the age-specific sensitivity to the common trend.
    \item $k_{t,i}$ is the country-specific mortality index, capturing local deviations.
    \item $b_{x,i}$ is the sensitivity of country $i$ to its own local deviation.
    \item $\varepsilon_{x,t,i} \sim \mathcal{N}(0, \sigma^2)$ represents the residual idiosyncratic noise.
\end{itemize}

The parameters are estimated through a two-step Singular Value Decomposition (SVD). First, the common factors ($B_x, K_t$) are extracted from the average mortality surface of the cluster. Subsequently, the residuals of this first decomposition are subjected to a second SVD to determine the country-specific components ($b_{x,i}, k_{t,i}$).

As a complementary benchmark for advanced ages, we also implemented the Cairns-Blake-Dowd (CBD) model \cite{cbd2006} for the age range 65--90, following the comparative framework established by Cairns et al. \cite{cairns2009}. The CBD analysis confirmed the presence of non-linear dynamics in the mortality curve rotation parameter, further supporting the case for a neural approach. While the CBD results are not the primary focus of this paper, they served as an important stress test during model development.

\subsection{The Forecasting Challenge}\label{sec:forecasting_challenge}
In the traditional framework, forecasting is achieved by modeling the mortality indices as stochastic processes. The common trend $K_t$ is typically assumed to follow a Random Walk with Drift:
\begin{equation}
    K_t = K_{t-1} + d + \xi_t
\end{equation}
where $d$ is the drift term and $\xi_t$ is white noise. Conversely, the local indices $k_{t,i}$ are required to be stationary, often modeled via an $AR(1)$ process as discussed in Section \ref{sec:stationarity_paradox}. 

\begin{figure}[!htbp]
    \centering
    \includegraphics[width=0.8\textwidth]{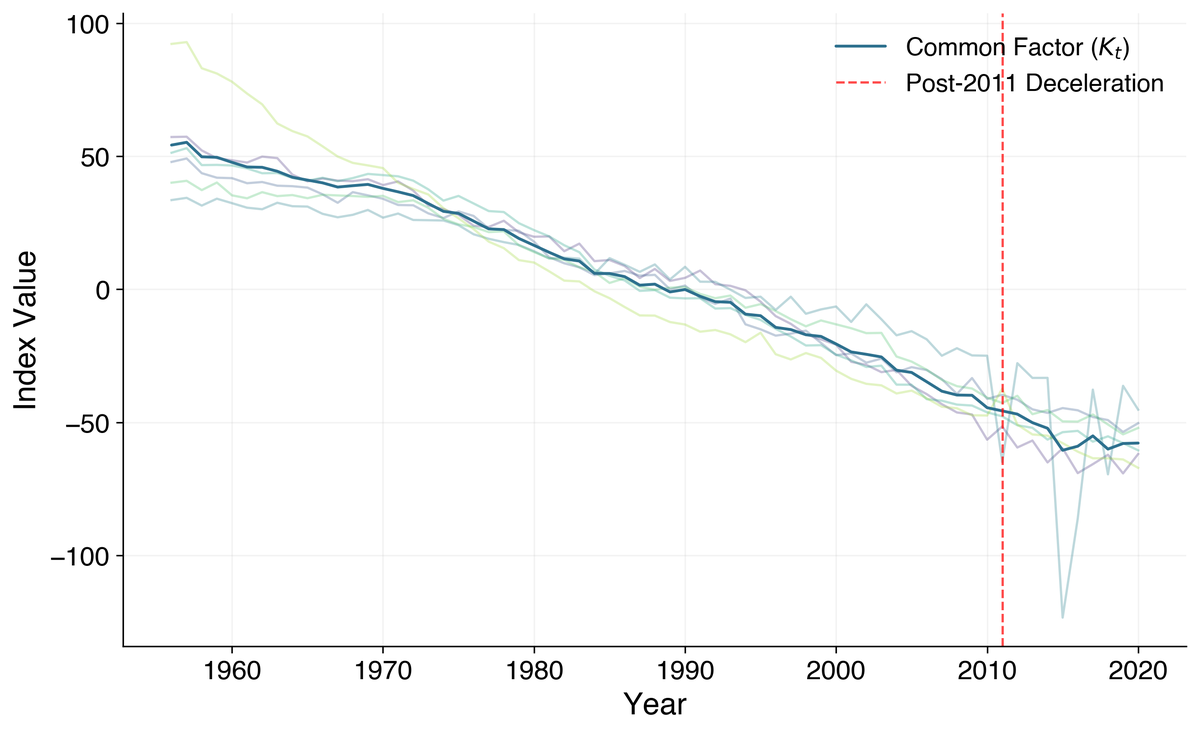}
    \caption{Evolution of the common mortality index $K_t$ for the frontier cluster. The red dashed line highlights the post-2011 deceleration, which represents a systemic departure from the historical linear drift.}
    \label{fig:common_kt}
\end{figure}

However, as illustrated in Figure \ref{fig:common_kt}, the common factor $K_t$ exhibits a distinct structural break around 2011. The linear drift $d$, estimated over the long-term horizon, fails to adapt to this recent deceleration in mortality improvement. 

Furthermore, the empirical failure of the mean-reversion assumption ($|\phi_i| \approx 1$) for the specific factors implies that the linear extrapolation of $k_{t,i}$ leads to significant projection bias. These dual limitations, namely the inability of the linear drift to capture the 2011 inflection and the persistence of country-specific deviations, motivate the transition toward a non-linear approach capable of internalizing complex temporal dependencies without the rigid constraints of structural stationarity.

\subsection{The Neural Transition and Feature Engineering}\label{sec:neural_transition}
To address the structural limitations of linear extrapolation and the persistent drifts identified in Section \ref{sec:stationarity_paradox}, we transition to a Deep Learning framework. The goal is to move from a rigid parametric process to a model capable of learning the underlying evolution operator of the mortality surface.

\subsubsection{First Differences and Data Stabilization}
As shown by the unit root analysis (Table \ref{tab:stationarity}), modeling absolute mortality levels $k_{t,i}$ leads to non-stationary sequences that are difficult for neural networks to generalize. In alignment with physics-informed approaches, we shift from a position-based representation to a velocity-based one. We define the state vector $\mathbf{V}_t$ at time $t$ as the set of first-order variations of the Li-Lee indices:
\begin{equation}
    \mathbf{V}_t = \left[ \Delta K_t, \Delta k_{t,1}, \dots, \Delta k_{t,N} \right]^\top \in \mathbb{R}^{N+1}
\end{equation}
where $\Delta Z_t = Z_t - Z_{t-1}$. Modeling variations instead of absolute levels serves a dual purpose: it stabilizes the mean of the series and prevents the network from simply memorizing historical levels, forcing it to learn the dynamics of mortality improvement.

\subsubsection{Sliding Window and Temporal Context}
Mortality is a path-dependent process where current improvements are influenced by historical trends. To provide the LSTM with sufficient context, we utilize a Sliding Window approach. The input to the network at time $t$ is a tensor $\mathcal{X}_t$ representing the sequence of the last $L$ variations:
\begin{equation}
    \mathcal{X}_t = \{ \mathbf{V}_{t-L}, \dots, \mathbf{V}_{t-1} \}
\end{equation}
We determined the lookback window $L$ through a sensitivity analysis across three horizons: 5, 10, and 15 years. While a 15-year window yields the lowest RMSE, we select $L=10$ years as the champion configuration. This selection optimizes the trade-off between data parsimony and model generalization, avoiding the risk of overfitting on old demographic regimes that are less relevant to current medical standards. This choice is further empirically justified by XAI validation (see Section \ref{sec:temporal_saliency}), which identifies a \textit{concentrated mid-horizon memory profile} with a dominant peak at lag $t-4$ (21.7\%), indicating that the model's predictive power is structurally rooted within this decadal memory depth.

\subsubsection{Feature Scaling and Anti-Leakage Protocol}
To ensure efficient convergence of the LSTM gradients, we apply a \texttt{StandardScaler} to normalize each component of $\mathbf{V}_t$ to zero mean and unit variance. To maintain the integrity of our out-of-sample validation, we implement a strict anti-leakage protocol: the scaling parameters are computed exclusively on the training set (1956--2011) and subsequently applied to the validation (2012--2020) and forecasting (2020--2050) periods without refitting. The train-validation split is strictly chronological, preserving the temporal ordering of the data. This ensures that no information about future drift magnitudes is leaked into the model during the training phase.

\subsection{The Hybrid-Lift Architecture}
Our proposed \textit{Hybrid-Lift} framework consists of a stacked LSTM architecture coupled with an explicit Mean-Bias Correction (MBC) layer. This hybrid approach allows the model to map the non-linear manifold of mortality variations while maintaining structural consistency with historical demographic trends.

\subsubsection{Stacked LSTM Core}
Temporal dynamics are processed through two stacked LSTM layers, whose configuration was determined via Bayesian hyperparameter optimization (15 trials, Keras Tuner). The first layer employs 32 hidden units, followed by a second layer of 16 hidden units that compresses the intermediate representation into a lower-dimensional feature space. We apply a fixed dropout rate of 20\% after the first LSTM layer, imposed as an architectural constraint to ensure the viability of Monte Carlo Dropout inference during the stochastic projection phase. This rate was not included in the hyperparameter search to guarantee that every candidate architecture supports Bayesian uncertainty quantification.

\subsubsection{Mean-Bias Correction (MBC) Adjustment}
A primary risk in utilizing recurrent architectures for long-term forecasting is neural drift, where compounding prediction errors lead to unrealistic mortality scenarios. To mitigate this, we introduce an MBC mechanism. We refine the raw LSTM output $\Delta \hat{\mathbf{V}}_t$ by applying a static bias correction based on the validation set, ensuring the long-term forecast is anchored to the observed historical reality:
\begin{equation}
    \hat{\mathbf{V}}_t = \Delta \hat{\mathbf{V}}_t + \mathcal{B}
\end{equation}
where $\Delta \hat{\mathbf{V}}_t$ represents the raw network prediction, and $\mathcal{B} \in \mathbb{R}^{N+1}$ is the empirical mean bias vector, computed as the average discrepancy between actual and predicted first-differences over the validation period (2012--2020):
\begin{equation}
    \mathcal{B} = \frac{1}{T_{\text{val}}} \sum_{t \in \mathcal{T}_{\text{val}}} \left( \mathbf{V}_t - \Delta \hat{\mathbf{V}}_t \right)
\end{equation}
where $\mathcal{T}_{\text{val}}$ denotes the set of validation time steps and $T_{\text{val}}$ its cardinality. This static, constant-offset adjustment forces the model to respect the historical drift of the cluster while allowing the LSTM to capture the non-linear fluctuations (the lift) around this baseline.

\subsubsection{Training Objective and Optimization}
The network is trained to minimize the Mean Squared Error (MSE) between predicted and actual first-differences:
\begin{equation}
    \mathcal{L}(\theta) = \frac{1}{T} \sum_{t=1}^T || \mathbf{V}_t - \hat{\mathbf{V}}_t(\theta) ||^2 
\end{equation}
where $T$ is the number of training samples and $\theta$ denotes the set of all trainable network weights. We utilize the Adam optimizer with an initial learning rate of $0.001$, implementing an Early Stopping criterion based on the validation set performance (with a patience of 15 epochs). This ensures the model generalizes to the frontier dynamics while preventing overfitting to the idiosyncratic noise inherent in the short-sample demographic regimes.

\subsection{Stochastic Projection and Uncertainty Quantification}

The inference framework combines two independent sources of uncertainty to generate stochastic mortality trajectories: \textit{model uncertainty}, captured via Monte Carlo Dropout, and \textit{process uncertainty}, calibrated on the historical variability of the Li-Lee benchmark.

For each time step of the recursive forecast (2020--2050), the LSTM champion model is invoked $S = 1{,}000$ times with active dropout layers, producing $S$ independent paths for each of the seven latent factors in the scaled space. Formally, the $s$-th stochastic prediction at time $t$ is:
\begin{equation}
    \hat{\mathbf{V}}_t^{(s)} = f_\theta(\mathcal{X}_t;\, \mathbf{z}^{(s)}), \quad s = 1, \dots, S
\end{equation}
where $f_\theta$ denotes the trained LSTM with frozen weights $\theta$, and $\mathbf{z}^{(s)}$ is the stochastic dropout mask sampled independently for each simulation. This component captures the \textit{epistemic uncertainty} arising from the stochastic regularization of the network.

After inverse-scaling to the original space, the level-reconstructed factors incorporate an additional process noise term:
\begin{equation}
    \hat{\mathbf{F}}_t^{(s)} = \hat{\mathbf{F}}_{t-1}^{(s)} + \hat{\mathbf{V}}_t^{(s)} + \boldsymbol{\eta}_t^{(s)}, \qquad \boldsymbol{\eta}_t^{(s)} \sim \mathcal{N}(\mathbf{0}, \text{diag}(\hat{\sigma}_1^2, \dots, \hat{\sigma}_{N+1}^2))
\end{equation}
where $\hat{\mathbf{F}}_t^{(s)}$ denotes the level-reconstructed factor vector at time $t$ for simulation $s$, and $\hat{\sigma}_j$ is the standard deviation of the historical first-differences of factor $j$, computed over the full 1956--2020 period (64 annual observations). This calibration anchors the process uncertainty to the intrinsic variability of the Li-Lee factor structure. The rationale is that the stochastic variability of the mortality process is a property of the demographic system, not of the forecasting model: regardless of whether one uses Li-Lee or an LSTM to predict the trend, the year-to-year fluctuations around that trend remain governed by the same underlying demographic dynamics.

The resulting ensemble $\{\hat{\mathbf{F}}_t^{(s)}\}_{s=1}^S$ thus reflects both the model's internal uncertainty (via dropout) and the irreducible variability of the mortality process (via calibrated noise), providing a more comprehensive basis for regulatory capital calculations than either component alone.

All quantitative results reported in this study, including mean projections, confidence intervals, SCR estimates, and SHAP attributions, are derived from a single realization of these trajectories and are fully reproducible under the fixed random seed provided in the public repository. Due to the stochastic nature of this inference framework, minor numerical variations are expected across independent runs with different seeds. However, qualitative findings, such as the relative ranking of model performance across countries and the convergence dynamics, remain robust across realizations.

\section{Results and Discussion}

\subsection{Forecasting Accuracy and MBC Impact}

To validate the predictive performance of the \textit{Hybrid-Lift} framework, we conducted an out-of-sample evaluation covering the period 2012--2020. This interval is critical, as it encompasses the post-2011 mortality deceleration noted in Section \ref{sec:forecasting_challenge}, providing a rigorous test for the LSTM's ability to adapt to structural breaks compared to the linear Li-Lee benchmark. We measure accuracy via Root Mean Squared Error (RMSE) on the level-reconstructed mortality indices, ensuring the Mean-Bias Correction (MBC) is applied to both models to maintain a fair comparison of structural adaptability. We note that this comparison evaluates the deterministic point forecasts of both frameworks (the Li-Lee Random Walk with Drift central projection against the Hybrid-Lift median trajectory): the performance differential thus reflects differences in structural adaptability to the post-2011 regime, not differences in uncertainty calibration.

\begin{figure}[!htbp]
\centering
\includegraphics[width=0.8\textwidth]{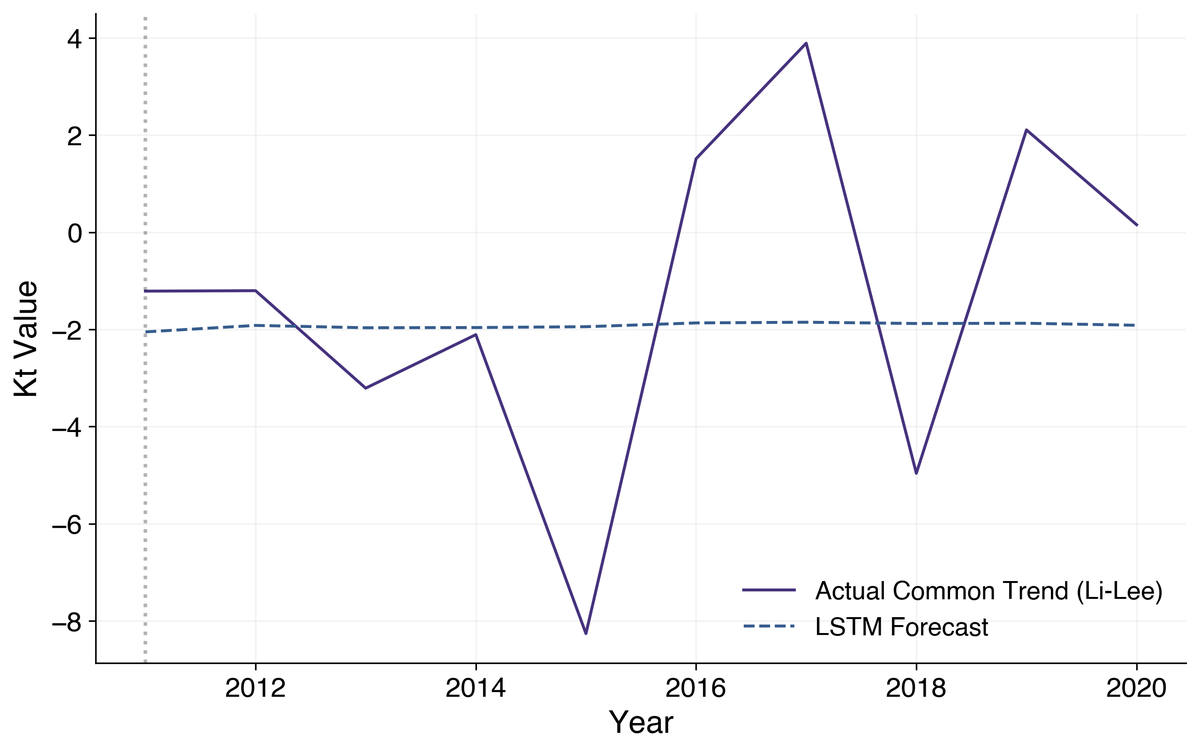}
\caption{Out-of-sample validation of the Hybrid-Lift mortality indices against observed values (2012--2020). The comparison illustrates the model's capacity to track the post-2011 structural deceleration in the common mortality trend.}
\label{fig:validation_kt}
\end{figure}

Table \ref{tab:benchmarking} summarizes the out-of-sample performance for the frontier cluster. The Hybrid-Lift framework shows consistent improvements in predictive accuracy across the majority of the analyzed nations, with particularly significant gains in regimes characterized by higher non-linear volatility, such as Sweden (17.40\% improvement) and West Germany (12.57\% improvement). Figure \ref{fig:validation_kt} provides a visual comparison of the predicted versus observed mortality indices over the validation period.

\begin{table}[!htbp]
\centering
\caption{Global Benchmarking Summary: Li-Lee vs. Hybrid-Lift (2012--2020)}
\label{tab:benchmarking}
\begin{tabular}{@{}lrrr@{}}
\toprule
Country & Li-Lee RMSE & Hybrid-Lift RMSE & Improvement (\%) \\ \midrule
Sweden & 2.7537 & 2.2744 & 17.403\% \\
West Germany & 0.9015 & 0.7882 & 12.570\% \\
Netherlands & 2.6024 & 2.4871 & 4.431\% \\
Norway & 7.9833 & 7.7026 & 3.516\% \\
Japan & 4.5646 & 4.6245 & -1.313\% \\
Switzerland & 1.3115 & 1.3387 & -2.078\% \\ \bottomrule
\end{tabular}
\end{table}

\subsubsection{Performance Analysis and Interpretation}

The results reveal a clear pattern: the Hybrid-Lift framework delivers its strongest improvements in populations where the stationarity paradox (Section \ref{sec:stationarity_paradox}) is most acute. Sweden, West Germany, and the Netherlands, the three countries where both ADF and KPSS tests confirm unit root behavior, exhibit RMSE improvements of 17.40\%, 12.57\%, and 4.43\%, respectively. Norway, which passes the stationarity tests but exhibits high local volatility, shows a moderate gain of 3.52\%. Conversely, Switzerland (-2.08\%) and Japan (-1.31\%), the two populations with near-linear historical trajectories and conflicting stationarity signals, show no benefit from the non-linear architecture. The absolute RMSE differences for these two countries are minimal (0.027 and 0.060, respectively), compared to improvements of 0.113 to 0.479 in the remaining four.

This pattern is not a limitation but a structural insight: a non-linear architecture offers no advantage when the underlying signal is itself well-approximated by a linear process. The Li-Lee model performs well for Switzerland and Japan precisely because their country-specific factors exhibit the near-stationary behavior that the model assumes. The value of Hybrid-Lift lies in its capacity to detect and exploit the persistent drifts that linear models are forced to ignore.

The Swiss case warrants specific discussion given its role as the primary case study for SST applications. The primary value of the Hybrid-Lift framework for Switzerland is not in outperforming Li-Lee on RMSE, where the underlying signal is near-linear, but in providing a governed neural overlay that integrates explainability (SHAP), uncertainty quantification (dual framework), and regulatory capital calibration (SCR) within a single auditable pipeline. In the context of internal model validation, the framework serves as a model challenger: it provides an independent, methodologically distinct projection that can be compared against the Li-Lee production model to identify potential blind spots or regime changes that the linear model might miss. The MBC anchoring mechanism ensures that this challenger remains conservative for stable trajectories, acting as a reliable guardrail rather than a disruptive engine that risks volatility in solvency buffers. Ablation studies (Section \ref{sec:ablation}) confirm that the framework's core design choices, namely first-differencing (+48.6\%) and Mean-Bias Correction (+18.6\%), contribute measurably to performance, even though the net improvement over Li-Lee for Switzerland is marginal.

\subsection{Stochastic Longevity Projections}\label{sec:stochastic_projections}

Beyond the prediction of latent mortality factors, the ultimate actuarial metric of interest is the impact on life expectancy at birth ($e_0$). The reconstruction pipeline proceeds in three stages: level integration, mortality surface recovery, and life table computation.

First, the level-reconstructed common factor $\hat{K}_t$ at time $t$ is obtained via cumulative integration of the bias-corrected first-differences:
\begin{equation}
    \hat{K}_t = K_{t_0} + \sum_{\tau=t_0+1}^{t} \left( \Delta \hat{K}_\tau + \mathcal{B}_0 \right)
\end{equation}
with $t_0$ being the last observed year and $\mathcal{B}_0$ the MBC component corresponding to the common factor.

Second, we map $\hat{K}_t$ back to log-mortality rates using a simplified Li-Lee reconstruction:
\begin{equation}
    \hat{y}_{x,t,i} = \alpha_{x,i} + B_x \hat{K}_t
\end{equation}
We employ this simplified reconstruction, which relies solely on the common factor $\hat{K}_t$, for two reasons. First, the country-specific dynamics are implicitly internalized by the LSTM during the joint training of all seven latent factors, meaning that the predicted $\hat{K}_t$ already reflects the co-evolution of common and specific components. Second, omitting the explicit specific terms $b_{x,i} k_{t,i}$ from the reconstruction avoids the accumulation of additional integration drift from six independent stochastic paths, yielding more stable long-term projections. A consequence of this choice is that cross-country dispersion in the reconstructed $e_0$ is driven primarily by differences in the baseline age profiles $\alpha_{x,i}$ rather than by divergent specific factors.

Third, from the reconstructed log-mortality surface, we derive period life tables using the standard actuarial approximation:
\begin{equation}
    q_x = \frac{m_x}{1 + 0.5\, m_x}, \qquad p_x = 1 - q_x, \qquad l_x = \prod_{j=0}^{x-1} p_j, \qquad e_0 = \sum_{x=0}^{\omega} l_x - 0.5
\end{equation}
where $m_x = \exp(\hat{y}_{x,t,i})$, $q_x$ is the probability of death, $p_x$ the survival probability, $l_x$ the survivorship function, and $\omega = 90$ is the terminal age. The subtraction of $0.5$ accounts for the standard mid-period mortality adjustment.

Table \ref{tab:longevity_summary} summarizes the projected mean gains in life expectancy for the frontier cluster over the 2020--2050 horizon, including the 95\% confidence intervals derived from the dual uncertainty framework.

\begin{figure}[!htbp]
\centering
\includegraphics[width=0.8\textwidth]{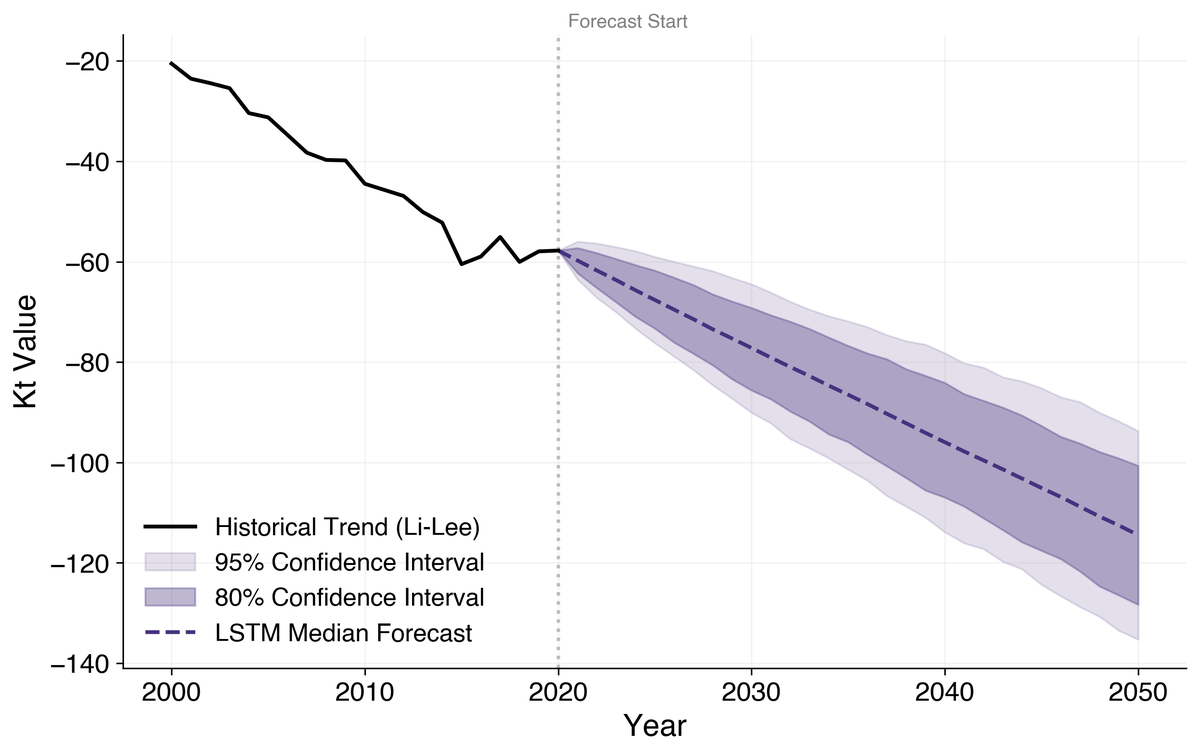}
\caption{Stochastic fan chart for the common mortality factor $K_t$ (2020--2050). The mean trajectory and 80\%/95\% confidence intervals are derived from 1,000 Monte Carlo Dropout realizations with Li-Lee calibrated process noise. The widening of the confidence bands over time reflects the accumulation of process uncertainty through the recursive forecast.}
\label{fig:kt_stochastic}
\end{figure}

Figure \ref{fig:kt_stochastic} presents the stochastic fan chart for the common mortality factor $K_t$, illustrating the projected trajectory and associated uncertainty bands over the 2020--2050 horizon.

\begin{table}[!htbp]
\centering
\caption{Stochastic Longevity Projections Summary (2020--2050)}
\label{tab:longevity_summary}
\begin{tabular}{@{}lrrrr@{}}
\toprule
Country & $e_0$ (2020) & $e_0$ (2050) & 95\% CI & Net Gain (Yrs) \\ \midrule
Switzerland & 81.71 & 84.77 & [83.80, 85.65] & +3.06 \\
Sweden & 81.71 & 84.73 & [83.76, 85.61] & +3.02 \\
Norway & 81.53 & 84.63 & [83.64, 85.52] & +3.09 \\
West Germany & 80.37 & 83.82 & [82.71, 84.83] & +3.45 \\
Netherlands & 81.30 & 84.46 & [83.45, 85.38] & +3.16 \\
Japan & 81.72 & 84.78 & [83.81, 85.66] & +3.06 \\ \bottomrule
\end{tabular}
\end{table}

\begin{figure}[!htbp]
\centering
\includegraphics[width=0.8\textwidth]{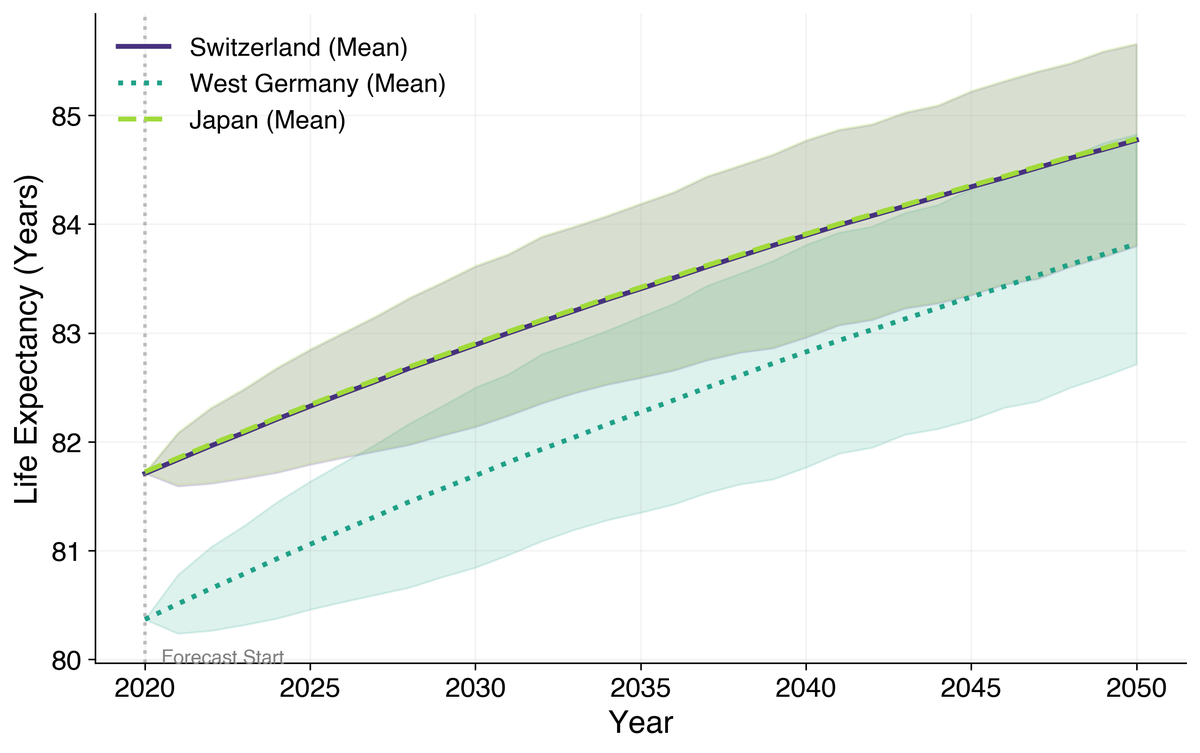}
\caption{Stochastic $e_0$ projections for selected frontier populations (2020--2050). Shaded areas represent the 95\% confidence intervals derived from the dual uncertainty framework (Monte Carlo Dropout and Li-Lee calibrated process noise). Note: Switzerland and Japan overlap almost entirely ($\Delta e_0 < 0.01$ years across the full horizon), reflecting their shared position on the biological longevity frontier.}
\label{fig:e0_forecast}
\end{figure}

The projections indicate a cluster-wide convergence toward a life expectancy of approximately 84.5--84.8 years by 2050. Notably, West Germany exhibits the most significant gain (+3.45 years), a result attributed to the model's capacity to internalize persistent local drifts that are often smoothed out in linear benchmarks. Figure \ref{fig:e0_forecast} illustrates the stochastic fan charts for the frontier leaders: Switzerland, Japan, and West Germany. A notable feature of these projections is the near-total overlap between Switzerland and Japan: their trajectories differ by only 0.01 years at both the 2020 origin ($e_0 = 81.71$ vs.\ 81.72) and the 2050 horizon (84.77 vs.\ 84.78), with virtually identical confidence intervals. This convergence is not a visualization artifact but a substantive demographic result: both nations occupy the same biological longevity frontier, and the model correctly identifies them as statistically indistinguishable under the dual uncertainty framework. The confidence intervals underscore the demographic coherence within this frontier cluster, even when subjected to the combined model and process uncertainty of the \textit{Hybrid-Lift} architecture.

\begin{figure}[!htbp]
\centering
\includegraphics[width=0.8\textwidth]{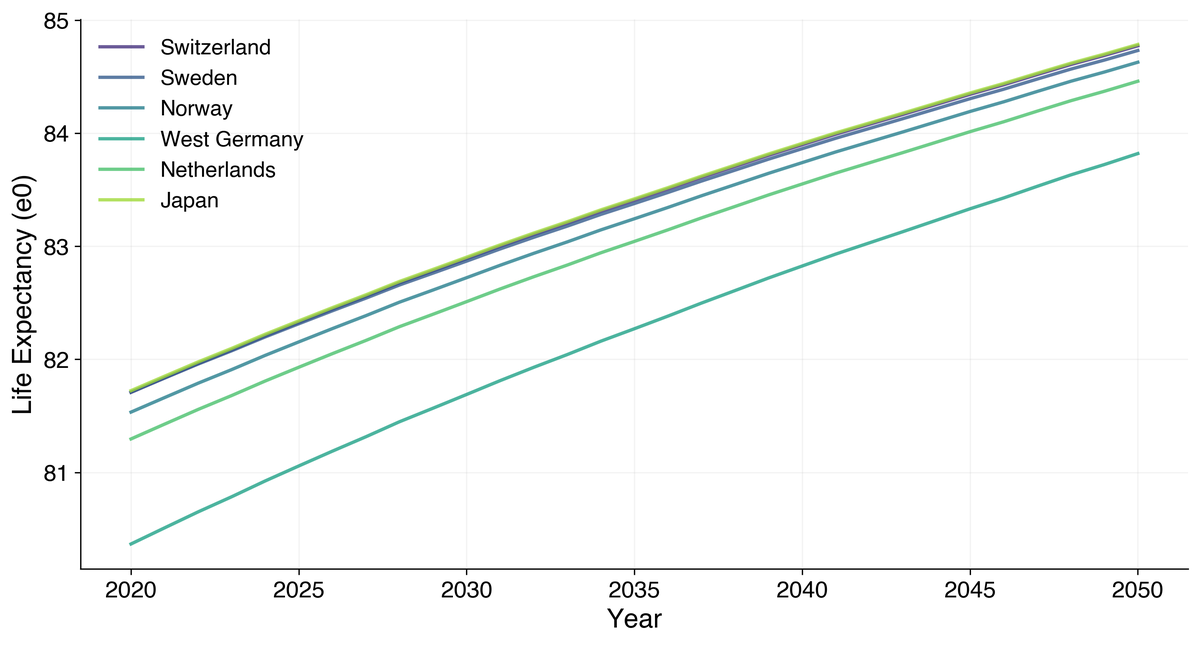}
\caption{Longevity convergence map for the full frontier cluster (2020--2050). Mean $e_0$ trajectories illustrate the systematic narrowing of cross-country differentials, with West Germany exhibiting the steepest improvement slope.}
\label{fig:convergence_map}
\end{figure}

Figure \ref{fig:convergence_map} visualizes the mean $e_0$ trajectories for all six nations, confirming the cluster-wide convergence dynamic and the catch-up effect exhibited by West Germany.

The confidence intervals observed in these stochastic projections reflect the dual uncertainty framework introduced above. The 95\% CI for Switzerland at the 2050 horizon spans approximately $\pm$0.9 years around the mean projection, a range that is consistent with the order of magnitude reported in long-term demographic projections by international agencies. To quantify the relative contribution of each uncertainty component: when only model uncertainty is active (MC Dropout without process noise), the 95\% CI contracts to approximately $\pm$0.01 years, two orders of magnitude smaller. This confirms that the process uncertainty component dominates the total uncertainty budget, accounting for over 99\% of the interval width.

This asymmetry is not a weakness of the framework but rather a structural insight with two distinct implications. First, the negligible contribution of MC Dropout to the total interval width indicates that the trained LSTM exhibits high internal consistency: the network's predictions are stable across stochastic dropout realizations, validating the architectural choices (layer sizes, dropout rate, training protocol) and confirming that the model has converged to a robust solution rather than an unstable local minimum. In this sense, MC Dropout serves primarily as a \textit{diagnostic of architectural stability}, i.e., a necessary validation step for regulatory acceptance of the neural component, rather than as the primary source of forecast uncertainty. The low epistemic variance is consistent with findings in low-dimensional time series forecasting tasks, where network capacity substantially exceeds the signal complexity, leading to well-determined weight configurations with minimal stochastic variation across dropout realizations. Second, the dominance of process uncertainty confirms that the meaningful variability in longevity projections arises from the intrinsic stochasticity of the demographic process itself, not from model indeterminacy. The calibration of this component on the Li-Lee historical variability ensures comparability with existing actuarial frameworks: the width of the confidence intervals is anchored to the same year-to-year fluctuations that underlie the industry-standard Random Walk with Drift projections. For regulatory capital calculations, this means the confidence intervals are driven by the same demographic fundamentals regardless of whether the trend is projected by Li-Lee or by the LSTM, ensuring that the neural model does not artificially inflate or deflate the solvency buffer relative to the established benchmark.

These projections indicate that while the common trend $K_t$ dictates the baseline longevity increase, the country-specific factors $k_{t,i}$ drive the cross-country divergence. The ability of our framework to produce distributional forecasts, rather than point estimates, is central to its utility in internal model validation and solvency assessment.

\section{Explainability and Model Validation}

\subsection{Temporal Saliency Analysis}\label{sec:temporal_saliency}

A fundamental requirement for the regulatory acceptance of neural models is the ability to explain \textit{what} the network has learned. To address this, we perform a gradient-based temporal saliency analysis that quantifies the relative importance of each time step within the $L=10$ input window. For a given input sequence $\mathcal{X}_t$, we compute the sensitivity of the predicted common factor variation $\Delta \hat{K}_t$ with respect to each lag:
\begin{equation}
    \mathcal{I}_\ell = \frac{1}{N+1} \sum_{j=1}^{N+1} \left| \frac{\partial \Delta \hat{K}_t}{\partial \mathcal{X}_{t,\ell,j}} \right|, \qquad \ell = 1, \dots, L
\end{equation}
where $\ell$ indexes the temporal position within the window (from $t-L$ to $t-1$) and $j$ indexes the feature dimension. The absolute gradient is averaged across all $N+1 = 7$ features to isolate the purely temporal contribution, and the resulting importance scores are normalized to sum to 100\%.

\begin{figure}[!htbp]
\centering
\includegraphics[width=0.8\textwidth]{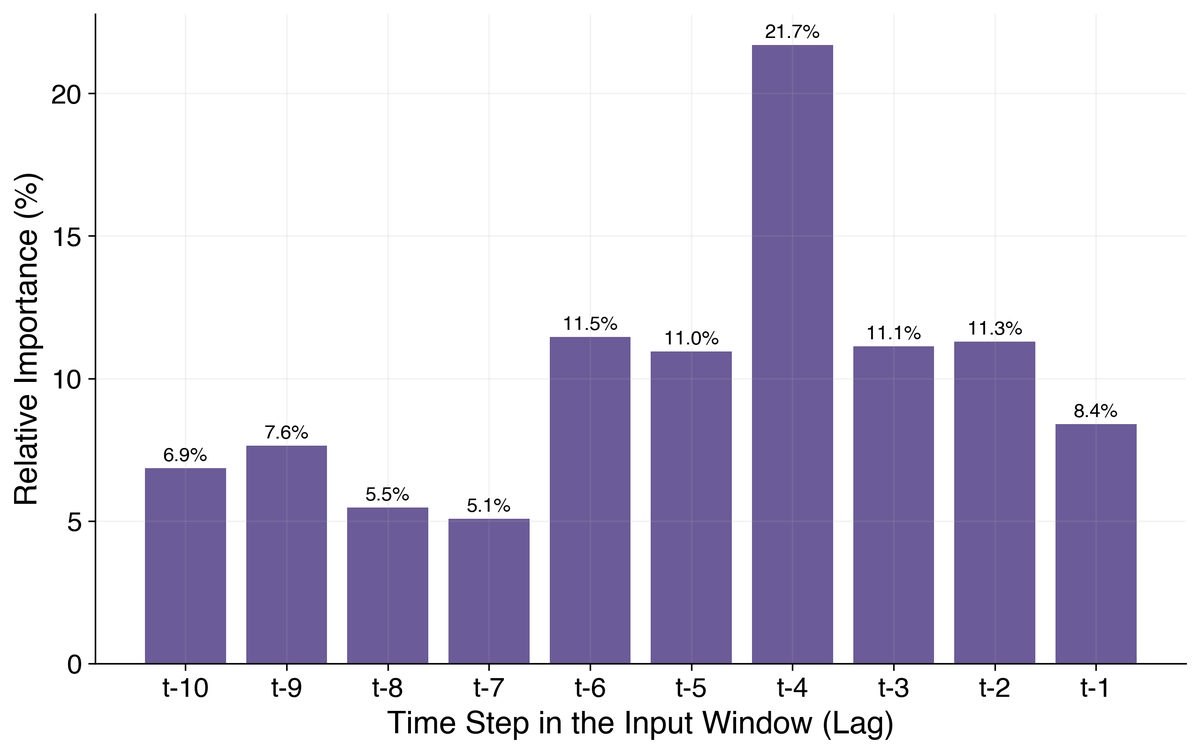}
\caption{Temporal saliency profile of the Hybrid-Lift LSTM. The dominant peak at $t-4$ (21.7\%) indicates concentrated sensitivity to medium-term structural shifts, while the non-zero importance at $t-10$ (6.9\%) validates the choice of a 10-year lookback window.}
\label{fig:temporal_importance}
\end{figure}

Figure \ref{fig:temporal_importance} presents the resulting memory profile. The analysis reveals a \textit{concentrated mid-horizon} pattern, with a dominant peak at lag $t-4$ (21.7\%), indicating that the LSTM assigns the highest predictive weight to observations approximately four years prior to the forecast origin. This is consistent with the model's sensitivity to the post-2011 mortality deceleration, which represents the most recent structural shift within the input window. Secondary importance is distributed across lags $t-6$ (11.5\%), $t-2$ (11.3\%), and $t-3$ (11.1\%), suggesting that the network integrates information from the full decadal horizon rather than relying exclusively on the most recent observation.

\begin{figure}[!htbp]
\centering
\includegraphics[width=0.8\textwidth]{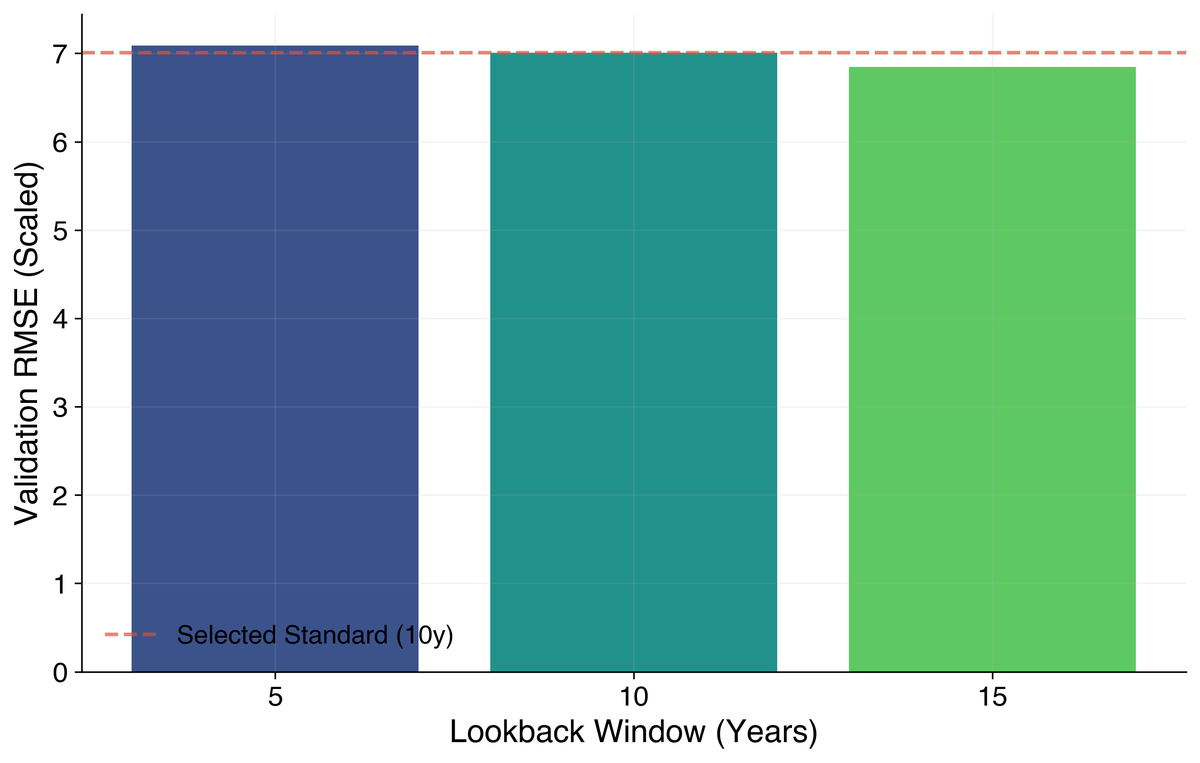}
\caption{Lookback window sensitivity analysis. RMSE decreases monotonically with window length, but the marginal gain from 10 to 15 years (+2.3\%) does not justify the reduction in training sample size.}
\label{fig:lookback_sensitivity}
\end{figure}

The non-zero importance at the window boundary ($t-10$: 6.9\%) provides empirical validation for the choice of $L=10$. To further substantiate this design decision, we conducted a sensitivity analysis across three lookback horizons (5, 10, and 15 years), retraining the champion architecture for each configuration. The results, illustrated in Figure \ref{fig:lookback_sensitivity}, confirm that RMSE decreases monotonically with window length: 7.09 (5 years), 7.01 (10 years), and 6.85 (15 years). While the 15-year window yields the lowest error, the marginal improvement over the 10-year configuration (+2.3\%) does not justify the loss of five additional training samples per country, a material reduction in a dataset spanning 65 years. The 10-year window thus represents the optimal trade-off between temporal depth and statistical volume.

\subsection{Cross-Country Influence Mapping (SHAP)}

To complement the temporal analysis of Section \ref{sec:temporal_saliency}, we investigate the \textit{spatial} dimension of the LSTM's learned representations: which countries in the cluster exert the strongest influence on Swiss mortality projections? We address this question using SHapley Additive exPlanations (SHAP) \cite{shapforactuaries2023}, a game-theoretic framework that assigns each input feature a fair contribution to the model's output.

We apply the KernelExplainer to the flattened input tensor ($L \times (N+1) = 70$ features), using the full training set as background and the full validation set (10 samples) as test instances. For each of the seven output dimensions, SHAP computes the marginal contribution of every input feature. To isolate the cross-country influence on Switzerland, we extract the SHAP values corresponding to the Swiss output index, reshape them into the original $(L, N+1)$ structure, and average the absolute values across both the temporal and sample dimensions. This yields a single importance score per input factor:
\begin{equation}
    \mathcal{S}_j = \frac{1}{|\mathcal{T}_{\text{test}}|} \sum_{t \in \mathcal{T}_{\text{test}}} \frac{1}{L} \sum_{\ell=1}^{L} |\phi_{t,\ell,j}^{\text{CHE}}|
\end{equation}
where $\phi_{t,\ell,j}^{\text{CHE}}$ denotes the SHAP value of feature $j$ at lag $\ell$ for the Swiss output, and $\mathcal{T}_{\text{test}}$ is the set of test samples.

\begin{figure}[!htbp]
\centering
\includegraphics[width=0.8\textwidth]{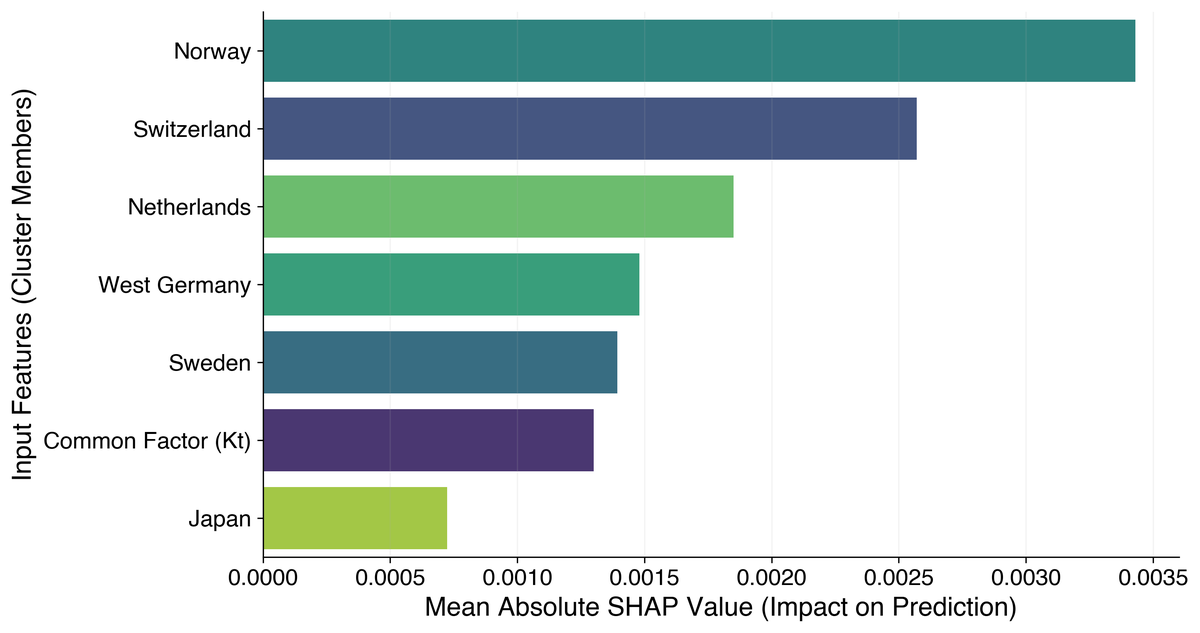}
\caption{SHAP influence mapping for Swiss mortality projections. The distributed importance across all cluster members indicates that the LSTM integrates information from the full multi-population structure rather than relying on a single dominant predictor.}
\label{fig:shap_influence}
\end{figure}

Figure \ref{fig:shap_influence} presents the resulting influence hierarchy for Switzerland. The analysis reveals a distributed pattern of cross-country contributions rather than a single dominant predictor. Norway emerges as the top external influence (0.00343), followed by Switzerland's own autoregressive signal (0.00257) and the Netherlands (0.00185). West Germany (0.00148), Sweden (0.00139), and the common factor $K_t$ (0.00130) contribute at comparable magnitudes, while Japan ranks lowest (0.00072), suggesting that the LSTM relies more on regional European signals than on the extra-European frontier for Swiss-specific predictions.

The SHAP values across countries are of the same order of magnitude, with the top-ranked feature (Norway, 0.00343) approximately 4.8 times larger than the lowest-ranked country (Japan, 0.00072), suggesting that the model exploits the full multi-population structure of the input rather than anchoring its predictions to a single lead-lag relationship. This finding is consistent with the design philosophy of the Hybrid-Lift framework, where the joint training of all seven latent factors encourages the network to internalize cross-country correlations. From a regulatory perspective, the absence of excessive dependence on any single external population is a desirable property, as it reduces the model's vulnerability to idiosyncratic shocks in individual countries.

\subsection{Biological Consistency via Li-Lee Age Profiles}\label{sec:biological_consistency}

A necessary condition for the regulatory acceptance of any mortality model is biological plausibility: projected death rates must increase monotonically with age in the adult range, consistent with the Gompertz law of exponential mortality growth. Violations of this property, i.e., cases where mortality at age $x+1$ falls below mortality at age $x$, would indicate that the model has learned spurious patterns incompatible with human biology.

To verify this, we reconstruct the full mortality curve $m_x$ for each country at the 2050 forecast horizon using the mean stochastic projections, and test whether $m_{x+1} \geq m_x$ holds for all ages in the range $x \in [30, 90]$. The lower bound of 30 is chosen to exclude the young-adult age group, where accident-related mortality can produce non-monotonic patterns that are biologically legitimate and unrelated to the aging process.

\begin{figure}[!htbp]
\centering
\includegraphics[width=0.8\textwidth]{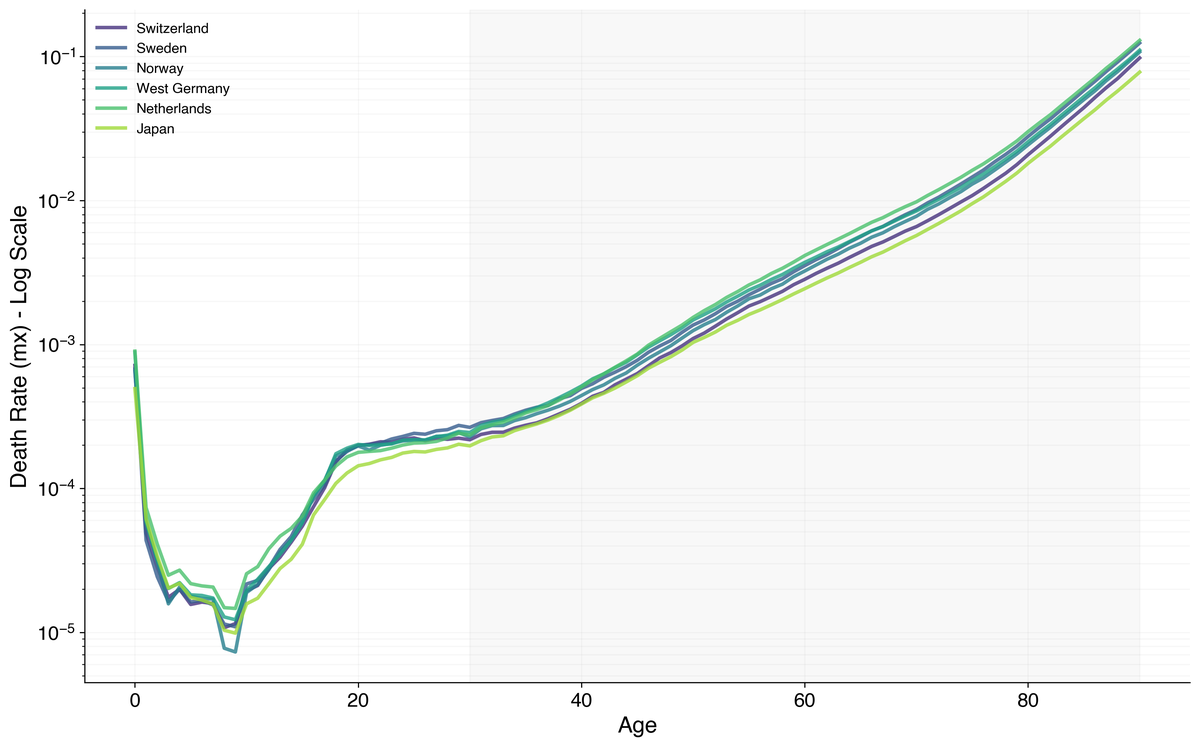}
\caption{Projected mortality curves for the frontier cluster at the 2050 horizon (log scale). All six populations pass the Gompertzian monotonicity test for ages 30--90, confirming biological consistency of the Hybrid-Lift projections.}
\label{fig:monotonicity}
\end{figure}

Figure \ref{fig:monotonicity} presents the projected mortality curves on a logarithmic scale. All six populations exhibit the characteristic Gompertzian exponential growth from age 30 onwards, with no crossings or inversions. The monotonicity test yields a PASS verdict for all countries in the cluster, confirming that the Hybrid-Lift framework preserves biological consistency despite the flexibility of the neural architecture.

This result warrants attention. Neural networks trained on first-differences operate in a transformed space where the original age structure is not explicitly enforced. The fact that the back-transformed mortality curves respect the Gompertz law without any explicit monotonicity constraint in the loss function suggests that the LSTM has implicitly learned the biological structure of the mortality surface through the Li-Lee age profiles $\alpha_{x,i}$ and $B_x$ used in the reconstruction step.

\subsection{Regulatory Capital and Scenario Analysis}\label{sec:regulatory_capital}

The capacity of a longevity model to inform regulatory capital calculations is its most direct measure of practical utility. We evaluate the Hybrid-Lift framework under two complementary perspectives: tail risk quantification for solvency purposes, and the impact of exogenous mortality shocks via scenario analysis.

\subsubsection{Solvency Capital Requirement}

We calibrate the Solvency Capital Requirement (SCR) using two standard tail risk measures: Value-at-Risk at the 99.5\% confidence level (VaR 99.5\%), as prescribed by Solvency II, and Expected Shortfall at the 99.0\% level (ES 99.0\%), as required by the Swiss Solvency Test (SST). For each country, the SCR is defined as the excess life expectancy beyond the mean projection at the 2050 horizon:
\begin{equation}
    \text{SCR}_{\text{VaR}} = \text{VaR}_{99.5\%}(e_0^{2050}) - \mathbb{E}[e_0^{2050}]
\end{equation}
\begin{equation}
    \text{SCR}_{\text{ES}} = \text{ES}_{99.0\%}(e_0^{2050}) - \mathbb{E}[e_0^{2050}]
\end{equation}
where the distributional quantities are derived from the 1,000 stochastic realizations of the dual uncertainty framework, and $\mathbb{E}[e_0^{2050}]$ denotes the mean (best estimate) of the projected life expectancy distribution.

\begin{figure}[!htbp]
\centering
\includegraphics[width=0.8\textwidth]{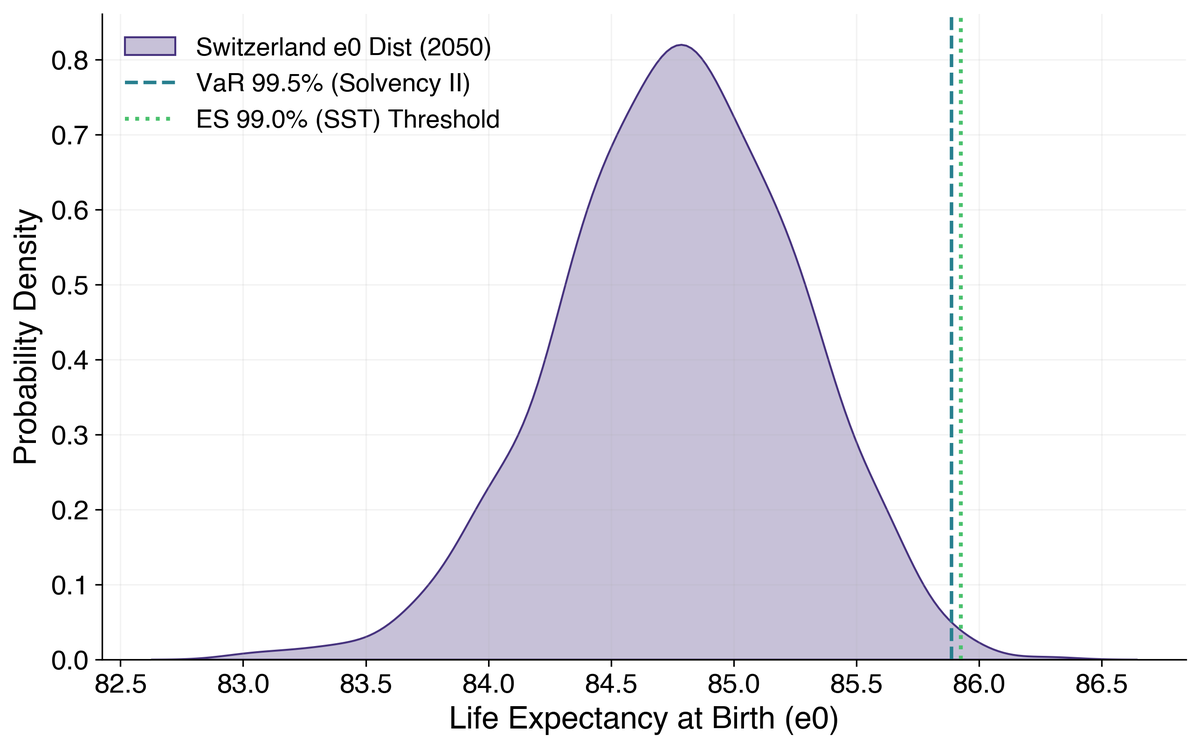}
\caption{Tail risk analysis for the Swiss $e_0$ distribution at the 2050 horizon. The VaR (99.5\%) and ES (99.0\%) thresholds are indicated, illustrating the risk margin derived from the dual uncertainty framework.}
\label{fig:tail_risk}
\end{figure}

Table \ref{tab:scr} presents the full cluster breakdown. Figure \ref{fig:tail_risk} illustrates the tail distribution for the Swiss case. The SCR values reflect the combined model and process uncertainty, with Switzerland, Japan, and Sweden exhibiting ES-based SCR values of approximately +1.15 years and West Germany showing a higher value of +1.32 years, consistent with its steeper catch-up trajectory. The gap between VaR and ES remains moderate across all countries, indicating well-behaved tail distributions without extreme fat-tail effects.

\begin{table}[!htbp]
\centering
\caption{Regulatory Capital Requirements: Full Cluster Breakdown (2050 Horizon)}
\label{tab:scr}
\begin{tabular}{@{}lrrrr@{}}
\toprule
Country & Mean $e_0$ & VaR 99.5\% & ES 99.0\% & SCR (ES) \\ \midrule
Switzerland & 84.77 & 85.89 & 85.93 & +1.153 yrs \\
Japan & 84.78 & 85.90 & 85.94 & +1.153 yrs \\
Sweden & 84.73 & 85.84 & 85.88 & +1.150 yrs \\
West Germany & 83.82 & 85.10 & 85.14 & +1.320 yrs \\
Netherlands & 84.46 & 85.62 & 85.66 & +1.203 yrs \\
Norway & 84.63 & 85.76 & 85.80 & +1.174 yrs \\ \bottomrule
\end{tabular}
\end{table}

\subsubsection{Reverse Stress Test}

A key requirement for internal model validation under the SST framework is the ability to identify the conditions under which the solvency buffer would be exhausted. Rather than applying arbitrary deterministic shocks and measuring their impact (an approach that introduces subjective assumptions about shock magnitude), we invert the problem: given the model's stochastic distribution, what is the critical mortality improvement that would consume the entire SCR?

Formally, we define the \textit{critical shock threshold} $\delta^*$ as the permanent, uniform reduction in age-specific mortality rates $m_x$ required to shift the mean projected life expectancy to the ES 99.0\% boundary:
\begin{equation}
    \delta^* : \quad \mathbb{E}[e_0^{2050} \mid m_x \to (1-\delta^*) m_x] = \text{ES}_{99.0\%}(e_0^{2050})
\end{equation}

The computation proceeds in three steps. First, we reconstruct the baseline Swiss mortality surface at the 2050 horizon using the mean common factor $\hat{K}_t$ from the 1,000 stochastic simulations and the Li-Lee age profiles: $\hat{m}_x = \exp(\alpha_{x,\text{CHE}} + B_x \cdot \bar{K}_{2050})$. Second, we apply uniform shocks at multiple levels (5\%, 10\%, 15\%, 20\%) and recompute $e_0$ from the shocked mortality surface via the standard actuarial life table. This yields a sensitivity coefficient $\partial e_0 / \partial \delta$, which we verify to be approximately constant across the tested range (coefficient of variation $< 1\%$), confirming the linearity of the $\delta \to \Delta e_0$ mapping for shocks up to 20\%. Third, we compute $\delta^* = \text{SCR}_{\text{ES}} / (\partial e_0 / \partial \delta)$.

For Switzerland, with a baseline mean $e_0$ of 84.77 years and an ES 99.0\% of 85.93 years (SCR = +1.153 years), the reverse stress test identifies a critical threshold of:
\begin{equation}
    \delta^* \approx 23.3\%
\end{equation}

This means that a permanent, uniform reduction in mortality rates of approximately 23\%, equivalent to a sustained, broad-spectrum medical breakthrough affecting all age groups simultaneously, would be required to exhaust the solvency buffer implied by the model's tail risk estimate. For context, a 10\% mortality reduction produces only $+0.494$ years of additional life expectancy, consuming approximately 43\% of the SCR buffer.

This result provides a direct, interpretable link between the model's statistical output and capital adequacy decisions. Risk managers can assess whether a shock of magnitude $\delta^*$ is plausible within the 30-year projection horizon, and calibrate their reserves accordingly. The threshold also serves as a benchmark for evaluating the severity of specific medical scenarios (e.g., cardiovascular breakthroughs, anti-aging therapies, or GLP-1 receptor agonist effects on all-cause mortality) against the model's implied risk tolerance.

It should be noted that the reverse stress test applies a deterministic shock to the reconstructed mortality surface rather than propagating it through the LSTM's input window. A natural extension would be to implement a full model-based stress test where shocked mortality rates are fed back into the network recursively, allowing the model's dynamic response, including potential non-linear amplification or dampening effects, to be evaluated directly.

\subsection{Ablation Studies}\label{sec:ablation}

To quantify the contribution of each key design choice, we conduct ablation experiments where individual components are removed from the Hybrid-Lift framework and the impact on out-of-sample RMSE is measured on the level-reconstructed common factor $K_t$ over the validation period (2012--2020).

\subsubsection{First Differences vs. Absolute Levels}
We retrain the champion architecture (32/16 units, 20\% dropout, $lr=0.001$) on absolute levels rather than first-differences, keeping all other settings identical. The RMSE increases from 15.90 (differences) to 30.93 (levels), a degradation of 48.6\%. This confirms that the first-difference transformation is essential for the network to generalize beyond the training period: without it, the model struggles to extrapolate the non-stationary trend of $K_t$.

\subsubsection{With vs. Without Mean-Bias Correction}
We reconstruct the validation-period levels from the champion's predicted differences without applying the MBC offset ($\mathcal{B} = 0$). The RMSE increases from 15.90 (with MBC) to 19.53 (without MBC), a degradation of 18.6\%. The MBC anchoring thus accounts for nearly one-fifth of the model's out-of-sample accuracy, confirming its role as a critical stabilizer against cumulative integration drift.

\begin{table}[!htbp]
\centering
\caption{Ablation Study: Impact of Key Design Choices on Out-of-Sample RMSE ($K_t$, 2012--2020)}
\label{tab:ablation}
\begin{tabular}{@{}lrr@{}}
\toprule
Configuration & RMSE ($K_t$) & Degradation \\ \midrule
Hybrid-Lift (Differences + MBC) & 15.90 & (baseline) \\
Without first-differences (levels) & 30.93 & +48.6\% \\
Without MBC & 19.53 & +18.6\% \\ \bottomrule
\end{tabular}
\end{table}

\section{Conclusion}

This paper introduced \textit{Hybrid-Lift}, a neural-actuarial framework for multi-population longevity forecasting that combines Hierarchical LSTM networks with Mean-Bias Correction anchoring. The framework was developed, validated, and applied to a frontier cluster of six high-longevity nations, with Switzerland as the primary case study for regulatory capital assessment under the Swiss Solvency Test. We position Hybrid-Lift as a governance-friendly model challenger, a neural overlay designed to complement, rather than replace, classical actuarial models by adding value where linear assumptions break down.

\subsection{Summary of Contributions}

The main contributions of this work are threefold. First, we identified a \textit{stationarity paradox} in the Li-Lee framework: for core European populations such as Sweden, West Germany, and the Netherlands, country-specific residuals exhibit persistent unit roots that violate the mean-reversion assumption underlying coherent multi-population forecasting. This finding provides a principled motivation for deploying a non-linear model challenger in parallel with the established linear benchmark.

Second, we proposed the Hybrid-Lift architecture, which addresses the limitations of both linear actuarial models and unconstrained neural networks. By modeling first-differences rather than absolute levels, fixing the dropout rate as an architectural constraint for Monte Carlo Dropout inference, and applying a static Mean-Bias Correction calibrated on the validation period, the framework achieves selective superiority: improvements of up to 17.40\% (Sweden) and 12.57\% (West Germany) in populations where unit roots invalidate the Li-Lee assumptions, while incurring a negligible performance cost ($<$ 2.1\%) for near-linear regimes such as Switzerland and Japan. This selective pattern is a structural feature, not a limitation: it confirms that the model adds value precisely where classical approaches are most vulnerable.

Third, we provided a comprehensive explainability and governance suite designed for internal model validation: gradient-based temporal saliency analysis revealing a concentrated mid-horizon memory profile; SHAP-based cross-country influence mapping showing distributed rather than concentrated dependencies; Gompertzian monotonicity compliance for all six populations; regulatory capital calibration under both Solvency II (VaR 99.5\%) and SST (ES 99.0\%) standards with a dual uncertainty framework; and a reverse stress test identifying the critical shock threshold ($\delta^* \approx 23.3\%$ for Switzerland) that would exhaust the solvency buffer. The MC Dropout component, while contributing negligibly to the total uncertainty budget, serves as a diagnostic of architectural stability, confirming that the trained network has converged to a robust solution suitable for production deployment.

\subsection{Limitations}

Several limitations should be acknowledged. The process uncertainty component of the confidence intervals is calibrated on the historical variability of the Li-Lee factor structure, which assumes that the stochastic properties of the mortality process remain stationary over the forecast horizon. If the demographic regime shifts fundamentally (e.g., through a medical breakthrough or a pandemic), the calibrated noise may underestimate or overestimate the true process variability. In particular, the Gaussian assumption underlying the process noise term may fail to capture the fat-tailed nature of extreme exogenous shocks, where the actual probability of large deviations exceeds what a normal distribution predicts. The reverse stress test partially addresses this concern by quantifying the critical shock magnitude, but a fully adaptive uncertainty calibration remains an open challenge.

The negligible contribution of MC Dropout to the total uncertainty budget raises a broader question about the role of epistemic uncertainty in neural mortality models. While we interpret this result as evidence of architectural stability, it also implies that the framework does not capture the uncertainty arising from model specification choices (e.g., alternative architectures, different lookback windows, or different training seeds). Robustness checks across random seeds and alternative train/validation splits would strengthen the confidence in the model's stability claims and are a natural extension of this work.

The SHAP-based influence hierarchy, while qualitatively informative, exhibits sensitivity to the model's trained weights and the inherent approximation of the KernelExplainer. The ranking of individual country contributions should be interpreted as indicative of the model's reliance on the full multi-population structure rather than as a stable quantitative ordering.

A methodological caveat concerns the Mean-Bias Correction: the bias vector $\mathcal{B}$ is calibrated on the same validation period (2012--2020) used to evaluate out-of-sample performance. While $\mathcal{B}$ is a simple constant offset (one scalar per feature) rather than a flexible model, this overlap means the reported RMSE improvements partially reflect the MBC's ability to correct for the specific drift observed in this particular validation window. With only 9 years of post-training data available, a separate held-out test set is impractical without further reducing an already small sample. This limitation should be considered when interpreting the absolute magnitude of the reported improvements.

Finally, the life expectancy reconstruction relies solely on the common factor $\hat{K}_t$ (via $\hat{y}_{x,t,i} = \alpha_{x,i} + B_x \hat{K}_t$), omitting the explicit country-specific terms $b_{x,i} k_{t,i}$. While this choice improves projection stability by avoiding the accumulation of six additional integration drift paths, it means that cross-country dispersion in the reconstructed $e_0$ is driven primarily by differences in the baseline age profiles $\alpha_{x,i}$. This simplification may understate the true cross-country divergence in scenarios where specific factors exhibit strong persistent trends, and partially limits the claim that country-specific dynamics are fully exploited in the final projections.

\subsection{Future Work}

This research opens several avenues for extension. A natural next step is the development of \textit{Actuarial-Informed Neural Networks}, analogous to Physics-Informed Neural Networks (PINNs), where actuarial constraints, such as Li-Lee coherence, Gompertzian monotonicity, and stationarity conditions, are embedded directly in the training loss function rather than verified post-hoc. A complementary direction is the application of credibility theory to the blending of neural and classical forecasts, where the relative weight assigned to each model is determined by a data-driven credibility factor. These extensions would strengthen the theoretical foundations of the neural-actuarial hybrid approach and further bridge the gap between algorithmic flexibility and actuarial governance. From a practical standpoint, the current framework models both sexes combined; extending it to sex-specific projections would be necessary for direct application to L\&H product portfolios (annuities, pension buy-outs) where male and female mortality dynamics diverge significantly. Beyond methodological advances, the stochastic trajectories produced by the framework can be directly applied to financial instruments such as longevity swaps, as illustrated in the accompanying codebase.

\section*{Acknowledgements}
The author thanks Riccardo Turin (Swiss Re, L\&H Risk Modeling) for valuable feedback on an earlier draft from the perspective of longevity risk management practice.

\newpage
\bibliographystyle{unsrt} 
\bibliography{references}  

\end{document}